\pdfoutput=1

\documentclass[11pt]{article}

\usepackage{emnlp2022}

\usepackage{times}
\usepackage{latexsym}
\usepackage{amsfonts}
\usepackage{xspace}
\usepackage{adjustbox}
\usepackage{multirow}

\usepackage{etoolbox}
\usepackage{siunitx,array}
\DeclareTextSymbol \textasteriskcentered{TS1}{42}
\newrobustcmd\B{\DeclareFontSeriesDefault[rm]{bf}{b}\bfseries}                          %
\newcommand{\prior}{\textasteriskcentered}

\usepackage[T1]{fontenc}

\usepackage[utf8]{inputenc}
\usepackage{csquotes}
\usepackage{microtype}
\usepackage{inconsolata}

\usepackage{soul}
\usepackage{xcolor}
\usepackage[normalem]{ulem}
\makeatletter
\newcommand{\rAB}[2]{\bgroup \UL@setULdepth
 \markoverwith{\lower\ULdepth\hbox
   {\kern-.03em\vbox{\color{#1}\hrule width.2em\kern1.2\p@\color{#2}\hrule}\kern-.03em}}%
 \ULon}
\makeatother
\newcommand{\summac}{\textsc{SC}\xspace}
\newcommand{\summacconv}{\summac{}\textsubscript{Conv}\xspace}
\newcommand{\summaczs}{\summac{}\textsubscript{ZS}\xspace}
\newcommand{\scuzs}{\textsc{SCU}\textsubscript{ZS}\xspace}
\newcommand{\sentli}{\textsc{seNtLI}\xspace}
\newcommand{\sentlirerank}{\textsc{seNtLI}\textsubscript{RR}\xspace}
\newcommand{\topk}{\textsc{TopK}\xspace}
\newcommand{\vitc}{\textsc{VitC}\xspace}
\newcommand{\roberta}{\textsc{Roberta}\textsubscript{MNLI}\xspace}
\newcommand{\ynie}{\textsc{Roberta}\textsubscript{ANLI}\xspace}

\title{Revisiting text decomposition methods for NLI-based factuality scoring of summaries}

\author{John Glover$^1$ \quad {\bf Federico Fancellu$^1$} \quad {\bf Vasudevan Jagannathan$^1$} \\ {\bf Matthew R. Gormley$^{1,2}$} \quad {\bf Thomas Schaaf$^1$} \AND
  {\normalfont $^1$3M Health Information Systems} \\ {\small \texttt{\{jglover,ffancellu,juggy,tschaaf\}@mmm.com}} \And
  {\normalfont $^2$Carnegie Mellon University}\\ {\small \texttt{mgormley@cs.cmu.edu}}}

\begin{document}
\maketitle

\begin{abstract}
Scoring the factuality of a generated summary involves measuring the degree to which a target text contains factual information using the input document as support.
Given the similarities in the problem formulation, previous work has shown that Natural Language Inference models can be effectively repurposed to perform this task.
As these models are trained to score entailment at a sentence level, several recent studies have shown that decomposing either the input document or the summary into sentences helps with factuality scoring.
But is fine-grained decomposition always a winning strategy?
In this paper we systematically compare different granularities of decomposition -- from document to sub-sentence level, and we show that the answer is no.
Our results show that incorporating additional context can yield improvement, but that this does not necessarily apply to all datasets.
We also show that small changes to previously proposed entailment-based scoring methods can result in better performance, highlighting the need for caution in model and methodology selection for downstream tasks.
\end{abstract}

\section{Introduction}
With improvements largely driven by recent advances in pre-trained language models~\citep{vaswani2017,radford2018,lewis2020b}, modern abstractive summarization models are capable of producing summaries that are both fluent and coherent.
However, they are still prone to various forms of ``hallucination'', generating statements that are not supported by the input text~\citep{cao2018,maynez2020}.
This has lead to a growing interest in being able to accurately measure the degree to which machine-generated output is non-factual~\citep{falke2019,kryscinski2020,pagnoni2021,laban2022}.

In factuality scoring and other closely related tasks such as fact verification~\citep{vlachos2014,thorne2018}, the objective is to assess whether or to what degree the claims in a given text can be supported by other ``evidence'' texts. 
Given this setup, previous work has drawn a parallel with the task of Natural Language Inference (NLI), which has a similar goal of determining whether the meaning of one text can be inferred (entailed) from another~\citep{dagan2006}.
As a consequence, models trained on large NLI datasets~\citep{bowman2015,williams2018,nie2020a} have often been successfully repurposed for the task of detecting factual inconsistencies in machine-generated summaries~\citep{falke2019,kryscinski2020,maynez2020,zhang2021d}.
It is now common that high-performance NLI models are trained on a combination of NLI and fact verification datasets~\citep{nie2020a,schuster2021}.

One way to repurpose NLI models for factuality scoring is to use the full text of the input and summary as the premise and hypothesis respectively, then take the factuality score to be a function of the model output distribution.
However, NLI models are usually trained with sentence pairs as input, and can suffer performance degradation with the longer contexts that arise in summarization~\citep{laban2022,honovich2022}.
Worse yet, the majority of modern NLI models are based on architectures such as the Transformer~\citep{vaswani2017} that use fixed-length input sizes, and it may not be possible for a full document and summary pair to fit into this context.

Another approach to NLI-based factuality scoring is grounded in the idea of first decomposing the input text into finer levels of granularity, followed by a later score aggregation step.
\citet{falke2019} proposed a scoring method based on sentence level decomposition, but concluded that the NLI models at the time were not robust enough for the task.
However, recently both \citet{schuster2022} and \citet{laban2022} have shown that variations on this decomposition-based strategy, in combination with the improved performance of modern NLI models, can produce systems that perform well at the task of detecting factual inconsistencies in generated summaries.

In this work we revisit existing studies of NLI-based factuality scoring and perform a systematic comparison of input-summary decomposition methodologies at different levels of granularity -- from document to sub-sentence level.
We show that contrary to previous findings, adding more context to the premise (the source document) can sometimes outperform approaches based on a more fine-grained decomposition.
We also find that small changes to the factuality scoring function can lead to a substantial increase in performance, but that model performance does not necessarily generalize across benchmarks that use different metrics (even when applied to the same underlying data).
Our results highlight the need for caution and additional evaluation when selecting a model and methodology for downstream tasks.

\section{Decomposition-based factuality scoring}
\label{sec:scoring-models}
In this work we are primarily concerned with \textit{referenceless} factuality scoring of document summaries.
To do so, we therefore require a function from an input $(document, summary)$ pair to a score value $Z \in \mathbb{R}$.
NLI models typically learn a function that maps a pair of input text strings $(X_{prem}, X_{hyp})$, commonly referred to as the \textit{premise} and \textit{hypothesis}, to a probability distribution over the output classes \textit{entailment}, \textit{neutral}, or \textit{contradiction}.
One simple way to repurpose NLI models for factuality scoring is with $(document, summary)$ as $(X_{prem}, X_{hyp})$, and to take the score $Z$ to be some function $f_Z(p_e, p_n, p_c)$ over the probability values given for entailment ($p_e$), neutral ($p_n$), or contradiction ($p_c$)\footnote{We note that generally the NLI models are not well-calibrated, and so these probability values may not necessarily have semantically meaningful interpretations, but empirically they can often be used directly in this manner.}.
We experiment with three decomposition-based scoring methods, described in the following sections.

\subsection{SummaC}
The SummaC models proposed by \citet{laban2022} decompose the document and summary into sentences.
A document is split into $M$ sentences labelled $D_1, \dots, D_M$, and a summary into $N$ sentences $S_1, \dots, S_N$.
Each $(D_m, S_n)$ combination is then passed through an NLI model, with scores computed using a function of the output probabilities.
This decomposition results in an $M \times N$ score matrix for each $(document, summary)$.
\citet{laban2022} describe two model classes, which differ in how they process the score matrix to create a final factuality score for a summary:\footnote{These models are agnostic to the particular NLI model being used for scoring, but the best performing model in the paper uses a version of ALBERT~\cite{lan2020a} fine-tuned on a combination of MNLI and VitaminC.}

\textsc{SummaC Zero-Shot} (\summaczs): each summary sentence is first scored by taking the maximum score value computed against any of the document sentences ($\max$ over each column in the $M \times N$ matrix). These summary sentence scores are then averaged to compute the final score.

\textsc{SummaC Convolution} (\summacconv): the pair matrix is converted to a histogram by placing the score values into evenly spaced bins, then the resulting matrix is passed through a 1-D convolutional layer. We refer the reader to \citet{laban2022} for further details.

We observe that although \citet{laban2022} indicate that the scoring function $f_Z$ that they use is given by $f_Z = p_e$, the default parameters in their publicly available code\footnote{\url{https://github.com/tingofurro/summac}} describe $f_Z = p_e - p_c$.
We compare these two variants of the score function $\mathbf f_Z$ in \S~\ref{sec:results}.

\subsection{\sentli}
Similarly to \citet{laban2022}, \citet{schuster2022} propose a factuality scoring model that assigns a score for each summary sentence $S_n$ according to the maximum score across all $(D_{1,\dots,M}, S_n)$ pairs.
Each $(D_m, S_n)$ is scored using a custom NLI model based on T5~\citep{2020t5} and fine-tuned on a combination of the SNLI~\citep{bowman2015}, MNLI~\citep{williams2018}, ANLI~\citep{nie2020a}, FEVER~\citep{thorne2018} and VitaminC~\citep{schuster2021} datasets.

Final scores are either the average score for all $S_{1,\dots,N}$ in an aggregation method referred to as ``soft aggregation'', or the \textit{minimum} score across $S_{1,\dots,N}$ in their ``hard aggregation'' method.
In addition, \citet{schuster2022} propose an extension to this approach called ``retrieve and rerank'' (\sentlirerank).
Here they again first score all $(D_m, S_n)$ using an NLI model.
For each $S_n$, the top-K $D_m$ are selected according to both the entailment and contradiction scores $p_e$ and $p_c$.
The NLI model is then presented with the same hypothesis $S_n$, together with a concatenation of the top-K entailing and contradicting sentences, with the output used to create the final score that $S_n$.
For further details we refer the reader to \citet{schuster2022}.

\subsection{Summarization Content Units (SCU)}
Following \citet{nenkova2004} and \citet{shapira2019}, we take decomposition a step further and segment each summary into smaller units called \textit{Summarization Content Units} (SCUs).
In its original formulation, SCUs are hand-crafted short spans of text describing a single fact contained in one or more reference summaries\footnote{Example SCUs are given in Appendix~\ref{sec:appendix-scus}}.
As our evaluation data is not manually annotated with SCUs, we follow the method in \citet{zhang2021d}, where the authors show that SCUs can be approximated using heuristics applied to the output of a Semantic Role Labeler.
However, whereas these methods apply to \textit{reference-based} evaluation of summaries, in the absence of human reference, here we adapt them to fit the reference\textit{less} evaluation scenario.
We refer to our method of decomposition and scoring with SCUs as \scuzs, and describe the details of the method in Appendix~\ref{sec:appendix-scus}.

\begin{table*}[t]
\fontsize{10.25pt}{12.3pt}\selectfont
\addtolength{\tabcolsep}{-0.75pt}
\begin{tabular}{lllllSSSSSS|S}
\cline{2-12}
\multirow{14}{*}{\rotatebox[origin=c]{90}{SummaC}} & System & $f_Z$ & PG & HG & {CGS~} & {XSF~} & {PT~} & {FCC~} & {SE~} & {FR~} & {Overall} \\
\cline{2-12}
& \summaczs & \multicolumn{3}{l}{}            & 70.4\prior & 58.4\prior & 62.0\prior & 83.8\prior & 78.7\prior & 79.0\prior & 72.1\prior\\
& \summacconv & \multicolumn{3}{l}{}          & 64.7\prior & 66.4\prior & \B 62.7\prior & 89.5\prior & 81.7\prior & 81.6\prior & 74.4\prior\\
& \sentli (soft) & \multicolumn{3}{l}{}       & 79.3\prior & 59.3\prior & 52.4\prior & 89.5\prior & 77.2\prior & 82.1\prior & 73.3\prior\\
& \sentlirerank (soft) & \multicolumn{3}{l}{} & 79.6\prior & 62.7\prior & 52.8\prior & 86.1\prior & 78.5\prior & 80.4\prior & 73.3\prior\\
& \sentlirerank (hard) & \multicolumn{3}{l}{} & \B 80.5\prior & 64.2\prior & 55.1\prior & 83.3\prior & 79.7\prior & 78.4\prior & 73.5\prior\\
\cline{2-12}
& \summaczs & $p_e - p_c$ & sent & sent & 62.5 & 53.8 & 57.6 & 83.9 & 77.1 & 79.2 & 69.0 \\
& \summaczs & $p_e$ & sent & sent       & 76.8 & 65.6 & 57.6 & \B 89.9 & 79.7 & 81.3 & 75.1 \\
& \summaczs & $p_e$ & doc & doc         & 59.3 & \B 69.9 & 59.9 & 84.7 & 78.7 & 81.2 & 72.3 \\
& \summaczs & $p_e$ & \topk & sent      & 79.7 & 67.3 & 56.9 & 89.4 & 81.8 & 81.4 & 76.1 \\
& \summaczs & $p_e - p_c$ & doc &  sent & 76.3 & 69.0 & 58.2 & 85.4 & 83.3 & \B 82.6 & 75.8 \\
& \summaczs & $p_e$ & doc &  sent       & 76.2 & 69.8 & 61.7 & 84.6 & \B 84.0 & 82.0 & \B 76.4 \\
& \scuzs & $p_e$ & \topk & SCU          & 72.9 & 65.6 & 57.1 & 80.5 & 82.1 & 81.7 & 73.3 \\
& \scuzs & $p_e$ & sent & SCU           & 71.4 & 63.4 & 55.0 & 77.0 & 80.0 & 81.4 & 71.4 \\ 
\multicolumn{11}{l}{} \\
\end{tabular}
\begin{tabular}{lllllS[table-format=1.2]S[table-format=1.2]S[table-format=1.2]S[table-format=1.2]}
\cline{2-9}
\multirow{11}{*}{\rotatebox[origin=c]{90}{FRANK}} & System & $f_Z$ & PG & HG & {Pearson $\rho$} & {\textit{p}-val} & {Spearman $r$} & {\textit{p}-val} \\
\cline{2-9}
& FactCC & & &          & 0.20\prior & 0.00\prior & 0.30\prior & 0.00\prior \\
& BertScore P Art & & & & 0.30\prior & 0.00\prior & 0.25\prior & 0.00\prior \\
\cline{2-9}
& \summaczs & $p_e - p_c$ & sent & sent  & 0.32 & 0.00 & 0.26 & 0.00 \\
& \summaczs & $p_e$       & sent & sent  & 0.35 & 0.00 & \B 0.36 & 0.00 \\
& \summaczs & $p_e$       & doc & doc    & 0.31 & 0.00 & 0.25 & 0.00 \\
& \summaczs & $p_e$       & \topk & sent & \B 0.37 & 0.00 & 0.34 & 0.00 \\
& \summaczs & $p_e - p_c$ & doc & sent   & 0.30 & 0.00 & 0.26 & 0.00 \\
& \summaczs & $p_e$       & doc & sent   & 0.34 & 0.00 & 0.29 & 0.00 \\
& \scuzs & $p_e$          & \topk & SCU  & 0.36 & 0.00 & 0.30 & 0.00 \\
& \scuzs & $p_e$          & sent & SCU   & 0.36 & 0.00 & 0.34 & 0.00 \\
\cline{2-9}
\end{tabular}
\caption{Test set results for SummaC and FRANK. Results marked ``\prior'' are taken from prior work, the rest are from our implementations. ``PG'' and  ``HG'' are the premise and hypothesis levels of granularity respectively. Sentences in our implementations are split using spaCy.}
\label{tab:test-results}
\end{table*}

\section{Experiments and evaluation}
\label{sec:experiments}
We evaluate the performance of our models on the SummaC benchmark~\citep{laban2022}, which comprises of six datasets for summary inconsistency detection: CoGenSumm (CGS)~\citep{falke2019}, XSumFaith (XSF)~\citep{maynez2020}, Polytope (PT)~\citep{huang2020a}, FactCC (FCC)~\citep{kryscinski2020}, SummEval (SE)~\citep{fabbri2021a}, and FRANK (FR)~\citep{pagnoni2021}.
Evaluation is standardized by casting each task as binary classification, and then measuring performance using balanced accuracy.
As the NLI-based factuality scoring methods all output a scalar score value, we follow \citet{laban2022} and tune thresholds separately for all methods and all datasets on the validation set, and report results using these threshold values on the test set.
Although the FRANK dataset is part of SummaC, we also perform a separate evaluation of it using the original metrics of Pearson and Spearman correlations of the model output scores with (non-binary) human scores.

To assess the benefits of decomposing text for NLI-based factuality scoring, we compare the performance of the aforementioned decomposition methods with full text scoring, where either or both the source document or the summary has not been decomposed.
We also test with a context length of several sentences, computed using a simplified version of the \sentlirerank method that we refer to as \topk, as follows:
\begin{itemize}
    \itemsep0em 
    \item First decompose the document and summary into individual sentences $(D_{1,\dots,M}, S_{1,\dots,N})$, and score all combinations using an NLI model.
    \item For each $S_n$, select the top-K sentences in $D_1,\dots,D_M$ according to $p_e$.
    \item Concatenate these top-K sentences to form a new premise string.
    \item Run hypothesis $S_n$ and the new premise through the NLI model, again taking $p_e$ as the final score for $S_n$.
    \item Compute the final factuality score as the average over the scores for each $S_n$.
\end{itemize}

To split text into sentences we use spaCy~\citep{honnibal2020}.
We note that \citet{laban2022} used NLTK~\citep{bird2009} for sentence-splitting, but this fails to correctly split sentences on some examples with bad punctuation (which are common in the FRANK dataset in particular\footnote{see Appendix~\ref{sec:appendix-sent-splitting} for details}).
In all experiments, unless otherwise specified we use the NLI model from \citet{schuster2021} that is fine-tuned on a combination of Vitamin-C and MNLI datasets\footnote{This is the best performing NLI model in \citet{laban2022}.}, which we refer to as \vitc.
For fair comparison with \citet{laban2022}, we set the maximum ``full document'' context for the premise to be 500 tokens.

\subsection{Results}
\label{sec:results}
Our main results are summarized in Table~\ref{tab:test-results}, with SummaC results at the top and FRANK results at the bottom.
In general, we find that factuality scoring using $f_Z = p_e$ has superior performance to $f_Z = p_e - p_c$, for all levels of input granularity, and for all evaluation metrics.
We surpass both the original \summaczs/\summacconv and \sentli/\sentlirerank SummaC results using \summaczs with this scoring function.
Further performance gains are also obtained from using additional context for the premise using \topk, and we find that including the full document context in the premise performs best of all, in contradiction to previous findings on this benchmark\footnote{In Appendix~\ref{sec:appendix-nli-models} we show that some of these findings appear to be unique to this particular choice of NLI model.}.
We see no additional performance benefit in going below the sentence level and using SCUs on these benchmarks, but the SCU decomposition does perform competitively across both benchmarks.

None of our variations achieve similar performance to the published \summaczs results, either performing better or worse depending on whether $f_Z$ is $p_e$ or $p_e - p_c$ respectively.
We believe that this discrepancy is due to the fact that the published \summaczs results use classification thresholds that are tuned on the test set\footnote{Confirmed via correspondence with \citet{laban2022}.} rather than validation set.

On FRANK, we find that there is no single method that performs best across both correlation metrics, \topk having the highest Pearson correlation, and the sentence level \summaczs the highest Spearman correlation.
It is notable however that the larger premise context granularity \textsc{doc-sent} is not as strong when using the original FRANK metrics as it is on SummaC, highlighting the need to be careful when comparing methods using different metrics, even on the same underlying data.

\section{Conclusion}
In this work we revisited prior findings that the best way to use NLI models for factuality scoring of machine-generated summaries is to first decompose the input to sentence level, score using NLI, then aggregate the sentence level scores to produce a document-level score.
Contrary to prior work, we find that there is no single optimal level of decomposition that performs best across all tasks and evaluation metrics.
We showed that in general, sentence level decomposition is preferable for the summary/hypothesis side of the NLI input, but on the premise side recent models such as \vitc often benefit from having longer input contexts available when scoring.
We also show that for the six datasets in the SummaC benchmark, there is still considerable variation in the performance of our methods both across the individual datasets, and also within different metrics on the same dataset.

\clearpage

\section*{Limitations}
Although we evaluate our methods across six different datasets, all are broadly from the same narrow domain, namely English news articles.
We also note that despite the methods in Section~\ref{sec:scoring-models} being agnostic to the choice of the NLI model that is used for scoring, there can be considerable degradation in the performance of methods that use longer premise contexts with some NLI models.
More details can be found in Appendix~\ref{sec:appendix-nli-models}.

\bibliography{custom}
\bibliographystyle{acl_natbib}

\appendix

\begin{table*}[t]
\begin{tabular}{llllcccccc|c}
\cline{2-11}
\multirow{5}{*}{\rotatebox[origin=c]{90}{SummaC}} & System & $f_Z$ & Splitter & CGS & XSF & PT & FCC & SE & FR & Overall \\
\cline{2-11}
& \summaczs & $p_e - p_c$ & NLTK  & 61.9 & 53.7 & 56.3 & 83.4 & 78.2 & 78.4 & 68.6 \\
& \summaczs & $p_e - p_c$ & spaCy & 62.5 & 53.8 & 57.6 & 83.9 & 77.1 & 79.2 & 69.0 \\
& \summaczs & $p_e$       & NLTK  & 75.6 & 65.3 & 60.4 & 89.5 & 80.1 & 79.1 & 75.0 \\
& \summaczs & $p_e$       & spaCy & 76.8 & 65.6 & 57.6 & 89.9 & 79.7 & 81.3 & 75.1 \\
\multicolumn{10}{l}{} \\
\end{tabular}
\begin{tabular}{llllcccc}
\cline{2-8}
\multirow{5}{*}{\rotatebox[origin=c]{90}{FRANK}} & System & $f_Z$ & Splitter & Pearson $\rho$ & \textit{p}-val & Spearman $r$ & \textit{p}-val \\
\cline{2-8}
& \summaczs & $p_e - p_c$ & spaCy & 0.27 & 0.00 & 0.23 & 0.00 \\
& \summaczs & $p_e - p_c$ & NLTK  & 0.32 & 0.00 & 0.26 & 0.00 \\
& \summaczs & $p_e$       & spaCy & 0.35 & 0.00 & 0.36 & 0.00 \\
& \summaczs & $p_e$       & NLTK  & 0.39 & 0.00 & 0.34 & 0.00 \\
\cline{2-8}
\end{tabular}
\caption{Performance differences on SummaC and FRANK test sets based on choice of sentence-splitting method. All methods use sentence level granularity for both premise and hypothesis. For SummaC all methods use thresholds selected using the validation set.}
\label{tab:results-sent-splitter}
\end{table*}

\begin{table*}[t]
\begin{tabular}{llllcccccc|c}
\cline{2-11}
\multirow{12}{*}{\rotatebox[origin=c]{90}{SummaC}} & System & PG & HG & CGS & XSF & PT & FCC & SE & FR & Overall \\
\cline{2-11}
& \roberta & doc & sent   & 58.1 & 56.2 & 52.9 & 62.5 & 57.0 & 66.2 & 58.8 \\
& \roberta & \topk & sent & 61.5 & 63.3 & 60.0 & 81.5 & 75.1 & 76.4 & 69.6 \\
& \roberta & sent &  sent & 75.2 & 61.3 & 59.2 & 90.7 & 80.1 & 79.5 & 74.3 \\
& \roberta & \topk & SCU  & 66.3 & 62.0 & 51.5 & 74.8 & 73.2 & 76.2 & 67.3 \\
& \roberta & sent & SCU   & 71.6 & 65.1 & 53.9 & 81.9 & 77.0 & 80.0 & 71.6 \\
\cline{2-11}
& \ynie & doc & sent      & 53.5 & 62.9 & 55.8 & 62.3 & 59.6 & 69.5 & 60.6 \\
& \ynie & \topk &  sent   & 77.3 & 65.4 & 58.4 & 82.4 & 78.4 & 76.9 & 73.2 \\
& \ynie & sent & sent     & 73.1 & 61.2 & 59.6 & 87.6 & 74.1 & 80.2 & 72.7 \\
& \ynie & \topk & SCU     & 74.5 & 64.3 & 59.2 & 82.1 & 77.8 & 77.9 & 72.6 \\
& \ynie & sent & SCU      & 70.8 & 64.4 & 55.8 & 79.8 & 76.7 & 81.0 & 71.4 \\
\multicolumn{10}{l}{} \\
\end{tabular}
\begin{tabular}{llllS[table-format=1.2]S[table-format=1.2]S[table-format=1.2]S[table-format=1.2]}
\cline{2-8}
\multirow{12}{*}{\rotatebox[origin=c]{90}{FRANK}} & System & PG & HG & {Pearson $\rho$} & {\textit{p}-val} & {Spearman $r$} & {\textit{p}-val} \\
\cline{2-8}
& \roberta & doc & sent   & 0.16 & 0.00 & 0.11 &  0.00 \\
& \roberta & \topk & sent & 0.23 & 0.00 & 0.21 &  0.00 \\
& \roberta & sent & sent  & 0.27 & 0.00 & 0.27 &  0.00 \\
& \roberta & \topk & SCU  & 0.23 & 0.00 & 0.21 &  0.00 \\
& \roberta & sent & SCU   & 0.26 & 0.00 & 0.27 &  0.00 \\
\cline{2-8}
& \ynie & doc & sent      & 0.05 & 0.04 & -0.04 & 0.16 \\
& \ynie & \topk & sent    & 0.27 & 0.00 & 0.30 &  0.00 \\
& \ynie & sent & sent     & 0.27 & 0.00 & 0.32 &  0.00 \\
& \ynie & \topk & SCU     & 0.24 & 0.00 & 0.23 &  0.00 \\
& \ynie & sent & SCU      & 0.27 & 0.00 & 0.29 &  0.00 \\
\cline{2-8}
\end{tabular}
\caption{Performance differences on SummaC and FRANK test sets based on choice of NLI model and level of granularity. For SummaC all methods use thresholds selected using the validation set. Sentences are split using spaCy. \roberta is the NLI model from \citet{liu2019c}, and \ynie is from \citet{nie2020a}.}
\label{tab:results-nli-models}
\end{table*}

\begin{table}[h]
\begin{tabular}{lSS}
\hline
\multicolumn{3}{c}{SummaC} \\
\hline
\multicolumn{3}{c}{} \\
& {NLTK} & {spaCy} \\
Mean         & 20.6 & 22.4 \\
Std. dev.    & 16.4 & 18.0 \\
$25^{th}$ \% & 11.0 & 12.0 \\
$50^{th}$ \% & 17.0 & 18.0 \\
$75^{th}$ \% & 26.0 & 28.0 \\
\multicolumn{3}{c}{} \\
\hline
\multicolumn{3}{c}{FRANK} \\
\hline
\multicolumn{3}{c}{} \\
& {NLTK} & {spaCy} \\
Mean         & 16.0 & 20.9 \\
Std. dev.    & 11.3 & 11.3 \\
$25^{th}$ \% & 7.0 & 13.0 \\
$50^{th}$ \% & 14.0 & 18.0 \\
$75^{th}$ \% & 24.0 & 28.0 \\
\end{tabular}
\caption{Mean, standard deviation, and percentiles of the number of sentences produced by NLTK and spaCy on SummaC and FRANK.}
\label{tab:sent-splitting}
\end{table}

\section{Performance variations with different sentence-splitting methods}
\label{sec:appendix-sent-splitting}
Table~\ref{tab:results-sent-splitter} describes how the performance of the SummaC Zero-Shot factuality scoring method varies based on whether NLTK or spaCy is used for sentence-splitting. All methods use the \vitc NLI model.
On SummaC, we see that using spaCy results in a slight improvement overall, whether our scoring function is $f_Z = pe$ or $f_Z = p_e - p_c$. We note that this is true for the FRANK dataset when scored using the SummaC balanced accuracy metric.
However, on the FRANK dataset with the original metrics, we mostly see the opposite effect; using NLTK results in higher Pearson correlations for both scoring functions, and a higher Spearman for $f_Z = p_e - p_c$.
Notably, the $0.39$ Pearson correlation for \summaczs at sentence level granularity using NLTK is the highest score that we obtain on this benchmark.

However, the results on Frank seem to be partly an artifact of inaccurate sentence-splitting by NLTK resulting in $(premise, hypotheis)$ pairs that are in fact at much larger levels of granularity than the intended sentence level, making this result difficult to interpret.
The following is an example of a passage of text taken verbatim from the FRANK validation set:
\begin{displayquote}
Thousands attended the early morning service at Hyde Park Corner and up to 400 people took part in a parade before the wreath-laying at the Cenotaph.Anzac Day commemorates the first major battle involving Australian and New Zealand forces during World War One.A service was also held at Westminster Abbey.The national anthems of New Zealand and Australia were sung as the service ended.
\end{displayquote}
\begin{figure}[h]
    \centering
    \includegraphics[width=\linewidth]{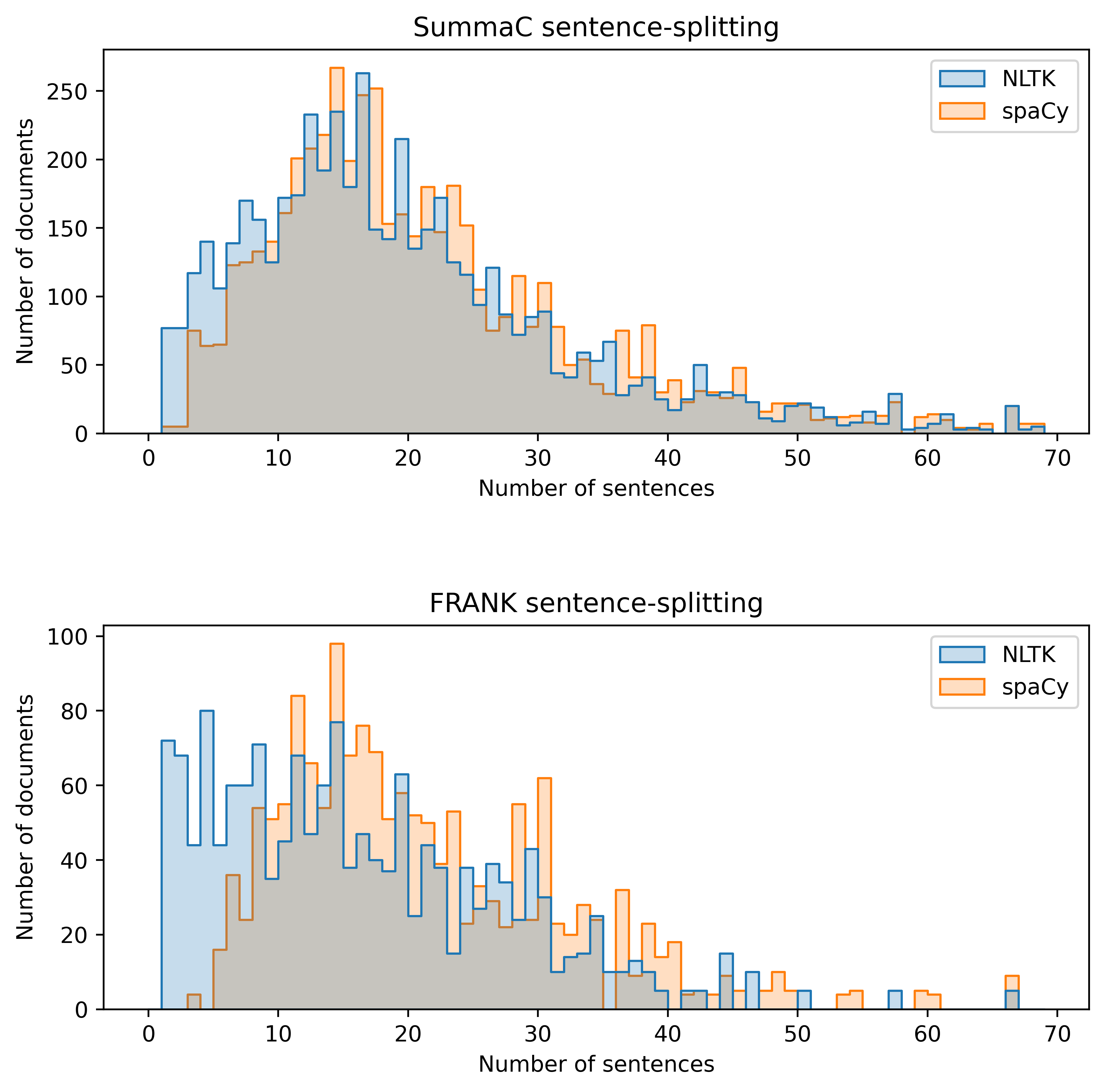}
    \caption{The number of sentences produced by NLTK and spaCy on SummaC and FRANK.}
    \label{fig:sent-splitting}
\end{figure}
We note that in this example there is no space after the fullstops, which causes NLTK's parser to break.
NLTK produces 1 sentence for this block of text, while spaCy produces 4 as we would expect.
This issue is relatively frequent in the FRANK dataset.
Figure~\ref{fig:sent-splitting} shows the distributions of the number of sentences produced by NLTK versus spaCy for all of the documents in both SummaC and FRANK, with statistics given in Table~\ref{tab:sent-splitting}.
We see that spaCy produces more sentences generally, with the difference being more pronounced on the FRANK dataset.

\section{Performance variations with different NLI models and levels of granularity}
\label{sec:appendix-nli-models}
In Table~\ref{tab:results-nli-models} we investigate how changing the level of decomposition effects the performance of two additional NLI models.
Notably with both of these models, scoring using the full document as the premise is significantly worse than either sentence level decomposition, \topk, or SCU, emphasizing that the results in Table~\ref{tab:results-nli-models} are highly dependent on the performance of the \vitc NLI model.
\topk and sentence level both perform reasonably well with these NLI models however, with the former being the best method to use on SummaC with \ynie and the latter the best with \roberta.
Again, we see no performance benefit when going to the SCU level.

\section{SCU examples}
\label{sec:appendix-scu-examples}
Two example one-line summaries, along with two extracted SCUs are shown below.
Colors indicate which parts of the generated summaries the SCUs are extracted from. 
\begin{itemize}
\item[] Summary$_1$: In 1998 \rAB{blue}{red}{two Libyans indicted} \setulcolor{red}\ul{in 1991} \rAB{blue}{red}{for the Lockerbie bombing} were still in Libya.
\item[] Summary$_2$: \rAB{blue}{red}{Two Libyans were indicted} \setulcolor{red}\ul{in 1991} \rAB{blue}{red}{for blowing up a Pan Am jumbo jet over Lockerbie}, Scotland in 1988.
\item[] SCUs: [\textcolor{blue}{two Libyans were officially accused of the Lockerbie bombing}, \textcolor{red}{the indictment of the two Lockerbie suspects was in 1991}]
\end{itemize}

\section{SCU-based decomposition details}
\label{sec:appendix-scus}
To create SCUs for a passage of text, we first split it into sentences using spaCy.
We then pass each sentence through co-reference resolution~\citep{joshi2020a}, and then finally we create SCUs using the method based on Semantic Role Labeling (SRL)~\citep{shi2019} described in \citet{zhang2021d}.
We use the publicly available code from \citet{zhang2021d}\footnote{\url{https://github.com/ZhangShiyue/Lite2-3Pyramid}, the authors refer to their SRL-based SCUs as \textit{Semantic Triplet Units} (STUs).} for both the co-reference resolution and SRL-based SCU generation.

To score a $(document, summary)$ pair, we experimented with decomposing either the document, the summary, or both into SCUs.
Here we describe the two variations that performed best on initial validation experiments.
The first scores summary SCUs against document sentences, and the second scores summary SCUs using longer passages of text from the document as context.

\subsection{\textsc{sent-SCU}}
This method is the most similar conceptually to \summaczs.
\begin{itemize}
  \itemsep0em 
  \item First decompose the document and summary into individual sentences $(D_{1,\dots,M}, S_{1,\dots,N})$, and then further decompose each $S_n$ into SCUs $S_{SCU_1},\dots,S_{SCU_J}$.
  \item Score all $(D_m, S_{SCU_j})$ combinations using an NLI model, and $f_Z = p_e$.
  \item The score for each $S_{SCU_j}$ is taken to be the maximum over the $(D_1,\dots,D_M, S_{SCU_j})$ pairs.
  \item For each $S_n$, average over the scores for $S_{SCU_1},\dots,S_{SCU_J}$ to calculate a score for that summary sentence, before averaging over the scores for each $S_n$ to create the document factuality score.
\end{itemize}

\subsection{\textsc{\topk-SCU}}
This is similar to the \topk scoring method from \S~\ref{sec:experiments}.
\begin{itemize}
  \itemsep0em 
  \item First decompose the document and summary into individual sentences $(D_{1,\dots,M}, S_{1,\dots,N})$, and then further decompose each $S_n$ into SCUs $S_{SCU_1},\dots,S_{SCU_J}$.
  \item Score all $(D_m, S_{SCU_j})$ combinations using an NLI model, and $f_Z = p_e$.
  \item For each $S_{SCU_j}$, we select the top-K sentences in $D_1,\dots,D_M$ according to $f_Z = p_e$, and concatenate them to form a new premise string.
  \item Hypothesis $S_{SCU_j}$ is re-scored using the new premise string, using $f_Z = p_e$ as the score for $S_{SCU_j}$.
  \item For each $S_n$ we then first average over the scores for $S_{SCU_1},\dots,S_{SCU_J}$ to calculate a score for that summary sentence, before averaging over the scores for each $S_n$ to create the document factuality score.
\end{itemize}

\end{document}